\documentclass{article}
\usepackage[utf8]{inputenc}

\usepackage{xcolor}
\usepackage{hyperref}
\usepackage{booktabs}
\usepackage{longtable}
\usepackage{amsfonts}
\usepackage{stfloats}
\usepackage{graphicx}
\usepackage[numbers, sort]{natbib}
\usepackage{authblk}
\usepackage{amsmath}
\usepackage{amssymb}
\usepackage{amsthm}
\usepackage[english]{babel}
\usepackage{charter}
\usepackage{fullpage}
\usepackage{pifont}
\usepackage{color,soul}
\usepackage{listings}
\usepackage{lipsum}
\usepackage{empheq}
\usepackage{subfigure}
\usepackage[most]{tcolorbox}
\usepackage{multirow}
\usepackage{tabularx}
\usepackage{makecell}
\usepackage{xspace} 
\usepackage{wrapfig}
\usepackage{caption}

\makeatletter
\def\thanks#1{\protected@xdef\@thanks{\@thanks
        \protect\footnotetext{#1}}}
\makeatother

\title{Computation-efficient Deep Learning for Computer Vision:\\A Survey}

\author{
   \vspace{0.5em}
   Yulin~Wang$^{\dagger*}$\thanks{$^*$Yulin~Wang and Yizeng~Han contribute equally to this work.\ \ \ \ $^\S$Corresponding author: Gao Huang.},\ 
   Yizeng~Han$^{\dagger*}$,\ 
   Chaofei~Wang$^{\dagger}$,\ 
   Shiji~Song$^\dagger$,\  
   Qi~Tian$^\ddagger$,\ 
   Gao~Huang$^{\dagger\S}$ \\
$^\dagger$Department of Automation, BNRist, Tsinghua University\ \ \ \ \ 
$^\ddagger$Huawei Inc. \\
\texttt{wang-yl19@mails.tsinghua.edu.cn\ \ \ gaohuang@tsinghua.edu.cn}\\
\vspace{-0.75em}
}

\date{}
\begin{document}

\maketitle

\renewcommand{\abstractname}{\large Abstract\\\vspace{0.5em}}

\begin{abstract}
	Over the past decade, deep learning models have exhibited considerable advancements, reaching or even exceeding human-level performance in a range of visual perception tasks. This remarkable progress has sparked interest in applying deep networks to real-world applications, such as autonomous vehicles, mobile devices, robotics, and edge computing. However, the challenge remains that state-of-the-art models usually demand significant computational resources, leading to impractical power consumption, latency, or carbon emissions in real-world scenarios. This trade-off between effectiveness and efficiency has catalyzed the emergence of a new research focus: computationally efficient deep learning, which strives to achieve satisfactory performance while minimizing the computational cost during inference. This review offers an extensive analysis of this rapidly evolving field by examining four key areas: 1) the development of static or dynamic light-weighted backbone models for the efficient extraction of discriminative deep representations; 2) the specialized network architectures or algorithms tailored for specific computer vision tasks; 3) the techniques employed for compressing deep learning models; and 4) the strategies for deploying efficient deep networks on hardware platforms. Additionally, we provide a systematic discussion on the critical challenges faced in this domain, such as network architecture design, training schemes, practical efficiency, and more realistic model compression approaches, as well as potential future research directions.
\end{abstract}

\section{Introduction}

Over the past decade, the field of computer vision has experienced significant advancements in deep learning. Innovations in model architectures and learning algorithms \cite{krizhevsky2017imagenet, simonyan2014very, szegedy2015going, He_2016_CVPR, huang2019convolutional, dosovitskiy2021image, liu2021swin} have allowed deep networks to approach or even exceed human-level performance on benchmark competition datasets for a wide range of visual tasks, such as image recognition \cite{dosovitskiy2021image, liu2021swin}, object detection \cite{zou2023object}, image segmentation \cite{minaee2021image, zhou2023survey}, video understanding \cite{kong2022human, sun2022human}, and 3D perception \cite{guo2020deep}. This considerable progress has stimulated interest in deploying deep models in practical applications, including self-driving cars, mobile devices, robotics, unmanned aerial vehicles, and internet of things devices \cite{chen2019deep, deng2019deep, chen2020deep}.


However, the demands of real-world applications are distinct from those of competitions. Models achieving state-of-the-art accuracy in competitions often exhibit computational intensity and resource requirements during inference. In contrast, computation is typically equivalent to practical latency, power consumption, and carbon emissions. Low-latency or real-time inference is crucial for ensuring security and enhancing user experience \cite{muhammad2020deep, bojarski2016end, grigorescu2020survey, zhang2019deep}. Deep learning systems must prioritize low power consumption to improve battery life or reduce energy costs \cite{mohammadi2018deep, li2018learning, li2018learning, carrio2017review}. Minimizing carbon emissions is also essential for environmental considerations \cite{caionce, xu2021survey}. Motivated by these practical challenges, a substantial portion of recent literature focuses on achieving a balance between effectiveness and computational efficiency. Ideally, deep learning models should yield accurate predictions while minimizing the computational cost during inference. This topic has given rise to numerous intriguing research questions and garnered significant attention from both academic and industrial sectors.

In light of these developments, this survey presents a comprehensive and systematic review of the exploration towards computationally efficient deep learning. Our aim is to provide an overview of this rapidly evolving field, summarize recent advances, and identify important challenges and potential directions for future research. Specifically, we will discuss existing works from the perspective of the following five directions:

\begin{figure*}[!t]
    \begin{center}
        \includegraphics[width=\linewidth]{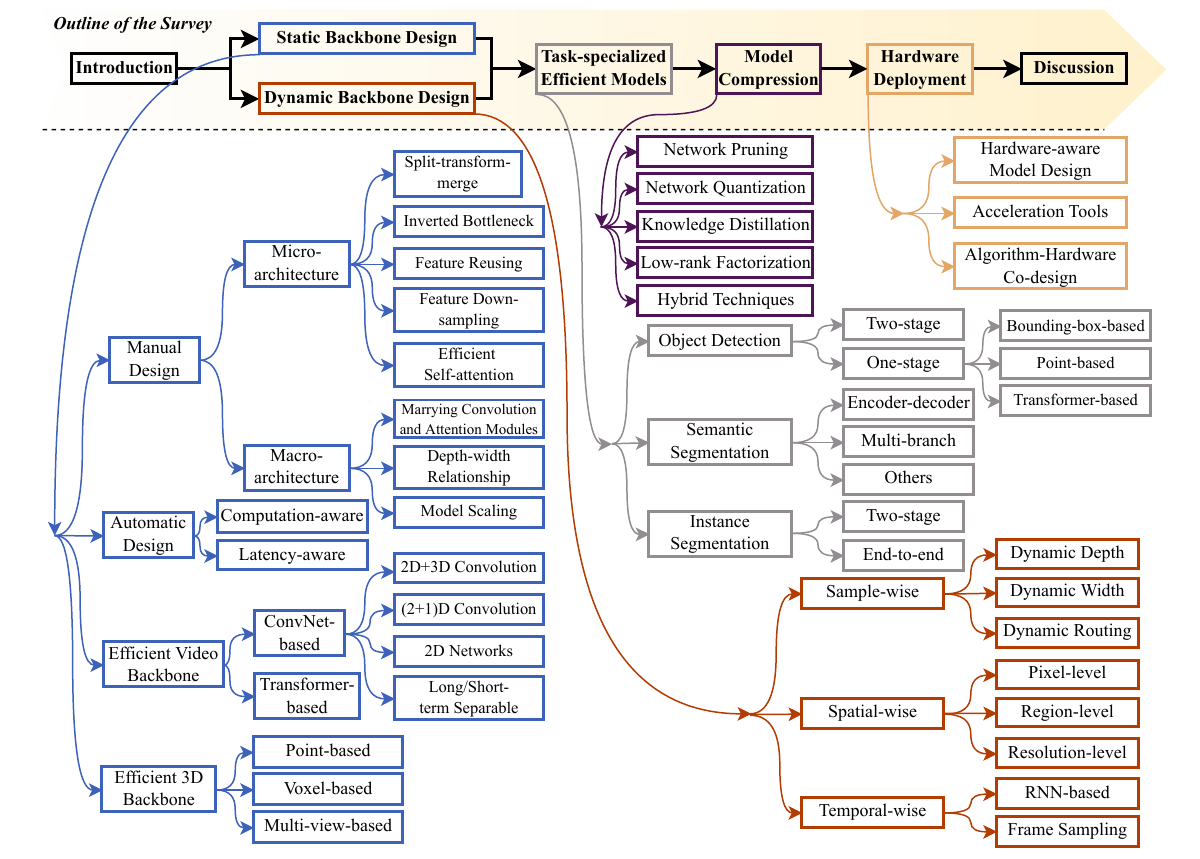}
    \end{center}
    \vskip -0.2in
    \captionsetup{font={small}}
    \caption{An overview of the survey. We first review the design of backbone networks, which are divided into static and dynamic models. Then we discuss the design of task-specialized algorithms and network architectures. Finally, we summarize the model compression approaches and the efficient hardware deployment techniques.}
    \label{fig:overview}
\end{figure*}

\textbf{1) Efficient Backbone Models.}
Designing light-weighted backbone networks that effectively extract discriminative deep representations from images, videos or 3D scenes with minimal computation by optimizing both efficient network micro-architectures (\emph{e.g.}, operators, modules, and layers) \cite{howard2017mobilenets, sandler2018mobilenetv2, howard2019searching} and improving the system-level organization of micro-architectures \cite{tan2019efficientnet, radosavovic2020designing}. Recent advances in neural architecture search (NAS) \cite{zophneural, liudarts} have further enabled the automatic design of backbones.

\textbf{2) Dynamic Deep Networks.}
Developing dynamic networks is an important emerging research direction for improving computational efficiency. These networks break the limits of static computational graphs and propose adapting their structures or parameters to the input during inference \cite{han2021dynamic}. For example, the model can selectively activate certain model components (\emph{e.g.}, layers \cite{huang2017multi}, channels \cite{lin2017runtime}, and sub-networks \cite{yan2015hd}) based on each test input or allocate less computation to less informative spatial/temporal regions \cite{wu2019adaframe, wang2021adaptive, hanlatency} of each input.

\textbf{3) Task-specialized Efficient Models.}
Numerous works focus on building task-specific heads on top of the features from light-weighted static/dynamic backbones to efficiently accomplish specific computer vision tasks. Examples include fast one-stage models for real-time object detection \cite{redmon2016you, liu2016ssd, tian2019fcos, wang2022yolov7}, the efficient multi-branch architecture for semantic segmentation \cite{yu2018bisenet}, and end-to-end instance segmentation frameworks \cite{wang2020solo, wang2020solov2}.

\textbf{4) Model Compression Techniques.}
Orthogonal to network architecture design, many algorithms have been proposed to compress relatively large models with minimal accuracy loss. This can be achieved by pruning less important network components \cite{liu2017learning, frankle2018the}, quantizing parameters \cite{courbariaux2015binaryconnect, rastegari2016xnor}, or distilling knowledge from large models to smaller models of interest \cite{hinton2015distilling, gou2021knowledge}.

\textbf{5) Efficient Deployment on Hardware.}
To achieve high practical efficiency, it is necessary to consider hardware requirements when developing deep learning applications. Reducing latency on specific hardware devices is usually treated as an objective in network design \cite{tan2019mnasnet, caiproxylessnas} or algorithm-hardware co-design \cite{hao2019fpga, jiang2020hardware}. Additionally, several acceleration tools have been developed for efficient deployment of deep learning models \cite{vanholder2016efficient, thakkar2019introduction, chen2018tvm}.

While some relevant surveys exist \cite{chen2019deep, chen2020deep_acm}, our survey is more up-to-date and comprehensive in several crucial aspects: 1) we systematically review model design techniques for images, videos, and 3D vision; 2) we summarize the recent works on designing dynamic deep neural networks for efficient inference; and 3) we thoroughly discuss the specialized models for accomplishing the most common and challenging computer vision tasks, \emph{e.g.}, object detection and image segmentation.


\begin{table*}[!t]
    \centering
    \begin{scriptsize}
    \captionsetup{font={small}}
    \caption{The ``\emph{split-transform-merge}'' architecture in representative computationally efficient deep networks. These blocks are typically adopted as basic components to build models. Here ``Conv'' refers to a convolutional layer.}
    \label{tab:tab1}
    \setlength{\tabcolsep}{0.5mm}{
    \renewcommand\arraystretch{1.49}
    \resizebox{\linewidth}{!}{
    \begin{tabular}{l|c|c|c|c|c|c|c|c}
    \hline
     & ResNet-50 & ResNeXt & Res2Net & MobileNet V2/V3 & EfficientNet & ShuffleNet & ShuffleNet V2 & Vision Transformer     \\[-0.9ex]
     & \cite{He_2016_CVPR} & \cite{xie2017aggregated}  & \cite{gao2019res2net} & \cite{sandler2018mobilenetv2, howard2019searching} & \cite{tan2019efficientnet} & \cite{zhang2018shufflenet}  & \cite{ma2018shufflenet} & \cite{dosovitskiy2021image, touvron2021training}     \\

    \hline
    \multirow{2}{*}{\emph{Split} function} & \multirow{2}{*}{1x1 Conv}  & \multirow{2}{*}{1x1 Conv}  & \multirow{2}{*}{1x1 Conv}  & \multirow{2}{*}{1x1 Conv}  & \multirow{2}{*}{1x1 Conv}  & \multirow{2}{*}{\shortstack{1x1 Group Conv\\+ Channel Shuffle}}  & \multirow{2}{*}{\shortstack{Channel Split\\+ 1x1 Conv}}  & \multirow{2}{*}{\shortstack{Linear Projection\\(\emph{i.e.}, 1x1 Conv)}}  \\
    &&&&&&&& \\
    \hline
    \multirow{2}{*}{\emph{Transform} function} & \multirow{2}{*}{3x3 Conv}  & \multirow{2}{*}{\shortstack{3x3 \\ Group Conv}}  &\multirow{2}{*}{\shortstack{Cascade\\3x3 Conv}}  & \multirow{2}{*}{\shortstack{3x3 or 5x5\\Depth-wise Conv}}  & \multirow{2}{*}{\shortstack{3x3 or 5x5\\Depth-wise Conv}}  & \multirow{2}{*}{\shortstack{3x3\\Depth-wise Conv}}  & \multirow{2}{*}{\shortstack{Identity and 3x3\\Depth-wise Conv}}  & \multirow{2}{*}{\shortstack{Multi-head\\Self-attention}}  \\
    &&&&&&&& \\
    \hline
    \multirow{2}{*}{\emph{Merge} function} & \multirow{2}{*}{1x1 Conv}  & \multirow{2}{*}{\shortstack{Concat\\+ 1x1 Conv}} & \multirow{2}{*}{\shortstack{Concat\\+ 1x1 Conv}} & \multirow{2}{*}{\shortstack{Concat\\+ 1x1 Conv}} & \multirow{2}{*}{\shortstack{Concat\\+ 1x1 Conv}} & \multirow{2}{*}{\shortstack{1x1 \\ Group Conv}}  & \multirow{2}{*}{\shortstack{Concat\\[-0.2ex] + 1x1 Conv + Concat\\[-0.2ex] + Channel Shuffle}}  & \multirow{2}{*}{\shortstack{Concat\\+ Linear Projection}} \\
    &&&&&&&& \\
    \hline

    \end{tabular}}}
    \end{scriptsize}
\end{table*}


The rest of this survey is organized as follows (see Figure \ref{fig:overview} for the overview). In Sec. \ref{sec:backbone} and \ref{sec:dy_backbone}, we introduce the design of efficient static and dynamic backbone networks, respectively. In Sec. \ref{sec:down_task}, the methodology for designing task-specialized efficient models is reviewed. The techniques for compressing deep learning models are investigated in Sec. \ref{sec:compress}. Efficient hardware deployment approaches are summarized in Sec. \ref{sec:hardware}. Lastly, we discuss existing challenges and future directions in Sec. \ref{sec:challenge}.

\section{Architecture Design of Backbone Networks}
\label{sec:backbone}

Typically, deep learning models for computation vision tasks incorporate two components, \emph{i.e.,} 1) a \emph{backbone network} that extracts deep representations from the raw inputs (\emph{e.g.}, images, video frames, and point clouds); and 2) a \emph{task-specific head} that is designed specialized for the task of interest. The deep features obtained from backbone networks are fed into the head to accomplish the corresponding task. The outputs of backbones (\emph{i.e.}, the inputs of the head) are usually assumed to have similar formats, while the outputs of the head are tailored for the tasks of interest.

In this section, we focus on how to design a computational-efficient general-purpose backbone network. Our discussions will start from processing the most fundamental data form, 2D images, where a light-weighted network may be obtained by either \emph{manual design} (Sec. \ref{sec:human_design}) or \emph{automatic searching approaches} (Sec. \ref{sec:nas}). Then we will discuss the backbones for processing \emph{videos} (Sec. \ref{sec:video_backbone}) and understanding \emph{3D scenes} (Sec. \ref{sec:3d_backbone}).

\subsection{Efficient Models by Manual Design}
\label{sec:human_design}

A considerable number of efficient backbone networks are designed manually based on theoretical derivations, empirical observations, or heuristics. Existing works can be categorized into two levels according the granularity of modifying the network: \emph{micro-architecture} (Sec. 2.1.1) and \emph{macro-architecture} (Sec. 2.1.2)

\subsubsection{Micro-architecture}

The micro-architecture refers to the individual layers, modules, and neural operators of backbones. These basic components are the foundation for constructing deep networks. Many works seek to attain higher computational efficiency by improving them. Notably, these works usually serve as off-the-shelf plug-in components that can be employed together with other techniques.

\textbf{1) Split-transform-merge Strategy.}
Typically, deep networks consist of multiple successively stacked layers with dense connection, where all the input neurons is connected to every output neuron. Formally, $\ell$-th layer $f^{\ell}$ with inputs $\mathbf{x}^{\ell-1}$ and outputs $\mathbf{x}^{\ell}$ can be expressed by
\begin{equation}
    \mathbf{x}^{\ell} = f^{\ell}(\mathbf{x}^{\ell-1}).
\end{equation}
However, such dense layers tend to computationally intensive. To address this issue, researchers have proposed to replace the dense connection with particularly designed topologies \cite{szegedy2015going, krizhevsky2017imagenet, xie2017aggregated}, which dramatically reduces the computational complexity, yet yields a competitive or stronger representation ability. Among existing works, one of the most popular designs is the \emph{split-transform-merge} strategy, as shown in the following (as a fundamental component, a residual connection \cite{He_2016_CVPR} is added here):
\begin{equation}
    \begin{split}
        \label{eq:split}
        &\{\mathbf{x}^{\ell-1}_1, \ldots, \mathbf{x}^{\ell-1}_C\} = f^{\ell}_{\textnormal{split}}(\mathbf{x}^{\ell-1}), \\
        \mathbf{x}^{\ell} &=\mathbf{x}^{\ell-1} + f^{\ell}_{\textnormal{merge}}(
            f^{\ell}_{1}(\mathbf{x}^{\ell-1}_1), \ldots, f^{\ell}_{C}(\mathbf{x}^{\ell-1}_C)
            ),
    \end{split}
\end{equation}
where $\mathbf{x}^{\ell-1}$ is \emph{split} into $C$ embeddings $\mathbf{x}^{\ell-1}_1, \ldots, \mathbf{x}^{\ell-1}_C$ with lower dimensions by a cheap operator $f^{\ell}_{\textnormal{split}}(\cdot)$. The low-dimensional embeddings are processed by the \emph{transform} functions $f^{\ell}_{1}(\cdot), \ldots, f^{\ell}_{C}(\cdot)$, whose input/output dimensions are the same. Notably, each of these functions corresponds to a dense layer, but this procedure is efficient due to the reduced input feature dimension. The processed embeddings are \emph{merged} by $f^{\ell}_{\textnormal{merge}}(\cdot)$. In the following, we will first discuss the design of the transform and split/merge functions respectively, and then introduce recent improvements over the ``split-transform-merge'' paradigm. Table \ref{tab:tab1} summarizes some representative \emph{split-transform-merge} architectures in popular computationally efficient deep networks.




\emph{a) ``Transform'' - homogeneous multi-branch architecture.}
A straightforward choice is to let $f^{\ell}_{c}(\cdot), c\!=\!1,\dots,C$ have the same architecture, and differentiate each other only in the values of the learnable parameters. This design is named as \emph{grouped convolution} \cite{krizhevsky2017imagenet, xie2017aggregated} when $f^{\ell}_{c}(\cdot)$ corresponds to a convolutional layer, and is adopted in many efficient ConvNets \cite{xie2017aggregated, li2019selective, zhang2022resnest}. IGCV \cite{zhang2017interleaved, xie2018interleaved, sun2018igcv3} further introduces a permutation operation to facilitate the interaction of different groups of embeddings (\emph{i.e.}, $\mathbf{x}^{\ell-1}_c, c\!=\!1,\dots,C$). In addition to convolution, this design is also widely adopted in the self-attention layers of vision Transformers (ViTs) \cite{dosovitskiy2021image, touvron2021training}. In ViTs, it is name as \emph{multi-head self-attention}, where $f^{\ell}_{c}(\cdot)$ is the scaled dot-product attention \cite{vaswani2017attention}.

In particular, the \emph{grouped convolution} is named as \emph{depth-wise separable convolution} when $C$ is equal to the channel number of $\mathbf{x}^{\ell-1}$. Here $f^{\ell}_{c}(\cdot)$ typically corresponds to a single convolution operation, where a large kernel size with sufficient receptive fields can be used without dramatically increasing the computational cost. This highly efficient component is first proposed in MobileNet \cite{howard2017mobilenets}, and adopted adopted in a wide variety of follow-up models \cite{chollet2017xception, sandler2018mobilenetv2, howard2019searching, mehta2019espnetv2, liu2022convnet, zhang2018shufflenet, ma2018shufflenet, tan2019efficientnet, wu2021cvt, guo2022cmt}.


\emph{b) ``Transform'' - heterogenous multi-branch architecture.}
Another line of works focus on developing nonequivalent branches, where each $f^{\ell}_{c}(\cdot), c\!=\!1,\dots,C$ is assigned with a specialized architecture or task. For example, the Inception architectures \cite{szegedy2015going, ioffe2015batch, szegedy2016rethinking, szegedy2017inception, si2022inception} adopt a varying receptive fields for different branches (\emph{e.g.}, by changing the convolution kernel size), aiming to aggregate the discriminative information at multiple levels. Recent works further extend this idea by feeding the outputs of $f^{\ell}_{c}(\cdot)$ to $f^{\ell}_{c+1}(\cdot)$ \cite{gao2019res2net, mehta2019espnetv2, raohornet}, and thus integrating multi-scale features into the outputs.

\emph{c) ``Split/merge'' functions}
$f^{\ell}_{\textnormal{split}}(\cdot)$ and $f^{\ell}_{\textnormal{merge}}(\cdot)$ are designed to map the features into or back from low-dimension embeddings with minimal cost. Most works adopt similar architectures for these two components: $f^{\ell}_{\textnormal{split}}(\cdot)$ corresponds to $1\!\times\!1$ convolution, while $f^{\ell}_{\textnormal{merge}}(\cdot)$ is accomplished by concatenation or concatenation + $1\!\times\!1$ convolution. Representative examples include ResNeXt \cite{xie2017aggregated}, MobileNets \cite{howard2017mobilenets, sandler2018mobilenetv2, howard2019searching} and Inception networks \cite{szegedy2015going, ioffe2015batch, szegedy2016rethinking, szegedy2017inception, si2022inception}. In particular, ShuffleNet \cite{zhang2018shufflenet} presents a more efficient design by combining $1\!\times\!1$ grouped convolution with channel shuffle.

\emph{d) Improved paradigms over ``split-transform-merge''.}
More recently, some works start to rethink the limitations of the split-transform-merge paradigm, and find that more efficient deep networks can be obtained by breaking this design principle. For example, motivated by the success of ViTs, ConvNeXt \cite{liu2022convnet} explicitly introduces an multilayer perceptron (MLP) at each layer by reverse $f^{\ell}_{\textnormal{split}}(\cdot)$ and the depth-wise separable convolution (\emph{i.e.}, $f^{\ell}_{c}(\cdot)$). EfficientNetV2 \cite{tan2021efficientnetv2} replace $f^{\ell}_{c}(\cdot)$ and $f^{\ell}_{\textnormal{split}}(\cdot)$ with a regular dense convolutional layer at earlier layers, achieving higher practical efficiency on GPU devices. MobileNeXt \cite{zhou2020rethinking} moves the depth-wise convolution layers to the two ends of the residual path to encode more expressive spatial information.



\textbf{2) Inverted Bottleneck.}
Bottleneck \cite{He_2016_CVPR} is a widely-used efficient component in ConvNets. Its basic architecture can be understood on top of Eq. (\ref{eq:split}): the total channel number of $\{\mathbf{x}^{\ell-1}_1, \ldots, \mathbf{x}^{\ell-1}_C\}$ will be reduced compared to $\mathbf{x}^{\ell-1}$ (\emph{e.g.}, by $4\times$ in ResNet \cite{He_2016_CVPR}). Consequently, the computationally intensive operations $f^{\ell}_{1}(\cdot), \ldots, f^{\ell}_{C}(\cdot)$ are performed on the low-dimensional embeddings, and the overall cost is saved. The effectiveness of this bottleneck is validated in both dense layers ($C\!=\!1$) \cite{He_2016_CVPR} and grouped convolution \cite{xie2017aggregated}. However, it may be sub-optimal in depth-wise separable convolution, where its low-dimensional transform results in information loss \cite{sandler2018mobilenetv2}. Inspired by this observation, MobileNetV2 \cite{sandler2018mobilenetv2} achieves an improved efficiency by proposing an inverted bottleneck, \emph{i.e.}, $\{\mathbf{x}^{\ell-1}_1, \ldots, \mathbf{x}^{\ell-1}_C\}$ have more dimensions than $\mathbf{x}^{\ell-1}$ (\emph{e.g.}, $6\times$ \cite{sandler2018mobilenetv2}). This designed is further adopted by a number of recent works \cite{sun2018igcv3, zhou2020rethinking, liu2022convnet}.


\textbf{3) Feature Reusing.}
Conventionally, the successive linear connection is the dominant topology for network design. The inputs are fed into a layer and transformed to obtained the inputs of the next layer. Any feature will be utilized for only a single time. Although being straightforward, this design is usually sub-optimal from the lens of computational efficiency. An important idea for lighted-weighted models is to \emph{reuse} the have-been-used features.

\emph{a) Inter-layer feature reusing.}
A basic idea is to reuse the features from previous layers. The skip-layer residual connection \cite{srivastava2015training, He_2016_CVPR} adds the inputs of each layer to the outputs, contributing the effective training of very deep and computationally more efficient networks. A more general formulation is established by dense connection \cite{huang2017densely, huang2019convolutional}, where all the previous features are fed into a next layer. CondenseNets \cite{huang2018condensenet, yang2021condensenet} extend this architecture by automatically learning the inter-layer connection topology. In contrast, other works like ShuffleNetV2 \cite{ma2018shufflenet} and G-GhostNet \cite{han2022ghostnets} focus on manually designing inter-layer interaction mechanisms.

\emph{b) Intra-layer feature reusing.}
The idea of feature reusing can also be leveraged within each network layers. For example, GhostNets \cite{han2020ghostnet, tangghostnetv2} demonstrate that there exist considerable redundancy in the outputs of each layer. They first obtain a small set of intrinsic output features, which are not only used as the inputs of the next layer, but also reused to generating other output features using cheap operations like linear transformations.


\textbf{4) Feature Down-sampling.}
Extracting deep representations from image-based data typically yields feature maps, which inherently have spatial sizes (\emph{i.e.}, height and weight). This property can be leveraged to reduce the computational cost of models \emph{e.g.}, introducing properly configured feature down-sampling modules.

\emph{a) Processing feature maps efficiently.}
The cost of processing feature maps grows quadratically with respect to their height/weight. OctConv \cite{chen2019drop} finds that processing all the features with the same resolution is not an optimal design. They propose to process a group of features at a down-sampled scale to capture only the low-frequency information, while the remaining features are designed to recognize high-frequency patterns, and the two groups exchange information after each layer. Consequently, the overall computational cost is reduced. This idea is also effective in ViTs \cite{panfast}. Similarly, HRNets \cite{sun2019deep, wang2020deep} and HRFormer \cite{yuan2021hrformer} maintain multi-resolution features at each layer, aiming to efficiently extract multi-scale discriminative representations for various computer vision tasks in the meantime.

\emph{b) Facilitating efficient self-attention.}
Particularly, feature down-sampling can be embedded into self-attention operations in ViTs to improve its efficiency. For example, PVTs \cite{wang2021pyramid, wang2022pvt} and ShuntedViT \cite{ren2022shunted} propose to compute attention maps efficiently with down-sampled feature maps. Twins \cite{chu2021twins} perform self-attention on low-resolution features to aggregate global information efficiently.

\textbf{5) Efficient Self-attention.}
ViTs \cite{dosovitskiy2021image} have achieved remarkable success in the fields of computer vision. Their self-attention mechanisms enable adaptively aggregating information across the entire image, yielding excellent scalability with the growing dataset scale or model size. However, vanilla self-attention suffers from high computational cost. A considerable number of recent visual backbones focus on developing more efficient self-attention modules without sacrificing their performance.

\emph{a) Locality-inspired Self-attention.}
In this direction, an important idea is drawn from the success of ConvNets: exploiting the locality of images, \emph{i.e.}, encouraging the models to aggregate more information from adjacent spatial regions. Swin Transformers \cite{liu2021swin, liu2022swin} achieve this by performing self-attention only within a square windows. Some other works extending this idea by designing different shapes of attention windows \cite{vaswani2021scaling, yang2021focal, dong2022cswin, yuan2022volo, xu2022v2x, tu2022maxvit} or introducing soft local constraints to attention maps \cite{d2021convit, li2022locality}. An important challenge faced by these works is how to model the interaction of different windows effectively. Possible solutions to address this issue include changing window positions \cite{liu2021swin, liu2022swin}, shuffling the channels \cite{fang2022msg}, designing specialized window shapes \cite{vaswani2021scaling, dong2022cswin, xu2022v2x, tu2022maxvit}, or further introducing window-level global self-attention modules \cite{yuan2021tokens, han2021transformer, chu2021twins}.

\emph{b) SoftMax-free Self-attention.}
To reduce the inherent high computation complexity of self-attention, another line of research proposes to replace the SoftMax function in self-attention with separate kernel functions, yielding linear attention \cite{katharopoulos2020transformers}. As representative examples, Performer \cite{choromanski2021rethinking} approximates SoftMax with orthogonal random features, while Nystr{\"o}mformer \cite{xiong2021nystromformer} and SOFT \cite{lu2021soft} attain this goal through matrix decomposition.  Castling-ViT \cite{you2022castling} measures the spectral similarity between tokens with linear angular kernels. EfficientViT \cite{cai2018efficient} further leverages depth-wise convolution to improve the local feature extraction ability of linear attention. FLatten Transformer proposes a focused linear attention module to achieve high expressiveness. \cite{han2023flatten}.

\subsubsection{Macro-architecture}

The macro-architecture refers to the system-level methodology of organizing micro-architectures (\emph{e.g.}, operators, modules and layers) and constructing the whole deep networks. Existing literature has revealed that, even with the same efficient micro-architectures, the approaches and configurations for combining them will significantly affect the computational efficiency of the resulting models. In the following, we will discuss the works and design principles relevant to this topic.

\textbf{1) Marrying Convolution and Attention Modules.}
Convolution and self-attention are both important modules with their own strengths. A considerable amount of literature has been published to study how to combine them for a higher overall computational efficiency. At the per-layer level, convolution can be leveraged to generate the inputs of self-attention, \emph{e.g.}, queries/keys/values \cite{wu2021cvt, guo2022cmt} or position embeddings \cite{xu2021co}. In addition, some works simultaneously utilize self-attention and a convolutional layer, and fuse their outputs \cite{lee2022mpvit, yu2021glance}, which facilitates the learning of local features. Another promising idea is to integrate convolution into the feed-forward network after the self-attention module \cite{yuan2021incorporating, li2021localvit, guo2022cmt}.

At the network level, many existing works focus on the placing order of self-attention and depth-wise convolution blocks. In particular, leveraging convolution at earlier layers is proven beneficial \cite{dai2021coatnet, xiao2021early, graham2021levit, mehtamobilevit, chen2021visformer}, which enables the efficient extraction of local representations. Besides, convolutional blocks are usually adopted as light-weighted down-sample layers \cite{mehtamobilevit, heo2021rethinking, dong2022cswin}. Another line of works parallelizes both a self-attention path and a convolution path in a single model \cite{xu2021vitae, zhang2023vitaev2, chen2022mobile, chen2022mixformer, peng2023conformer, peng2021conformer}, where the two paths typically interact in a layer-wise fashion.



\textbf{2) Depth-width Relationship.}
In the context of ConvNets and hierarchical ViTs, the backbone models consist of multiple stages with progressively reduced feature resolution. The layers within each stage usually have the same width, while later stages are wider. The stage-wise width growing rule is an important configuration, where it is popular to adopt an exponential growth with base two \cite{He_2016_CVPR, huang2019convolutional, liu2021swin}. In contrast, RegNets \cite{radosavovic2019network, radosavovic2020designing} further propose a more detailed principle: widths and depths of good networks can be explained by a quantized linear function.


\textbf{3) Model Scaling.}
On top of designing a single efficient model, it is also important to obtain a family of models that can adapt to varying computational budgets. An important principle for addressing this issue is \emph{compound scaling} \cite{tan2019efficientnet, tan2021efficientnetv2}, which indicates that simultaneously increasing the depth, width and input resolution of a given base model will yield a family of efficient network architectures. Doll{\'a}r \emph{et al.} \cite{dollar2021fast} further study how to design a proper model scaling rule in terms of the actual runtime. In addition, TinyNets \cite{han2020model} extend this idea to the shrinking of the model size.


\subsection{Automatic Architecture Design}
\label{sec:nas}
Compared to manually designing backbones, another appealing idea is to find proper network architectures automatically, which is usually referred to as \emph{neural architecture search (NAS)}. In recent years, a number of existing works have investigated this idea through the lens of computational efficiency. In the following, we will discuss the basic computation-aware formulation of NAS (Sec. 2.2.1)  and how the practical speed is considered in NAS  (Sec. 2.2.2).

\subsubsection{Computation-aware NAS}
Typically, NAS consists of two components: a searching space that contain a number of candidate architectures, and an algorithm to search for an optimal architecture. The computational cost for inferring the model is usually treated as a constraint, which is either inherently controlled by the searching space or strictly restricted by a pre-defined rule. The optimization objective is to maximize the validation accuracy.

\textbf{1) Early Works.}
Early NAS methods propose to formulate a discrete searching space \cite{zophneural, bakerdesigning, zoph2018learning}. The network is viewed as a graph with a number of nodes connected by edges, where each edge corresponds to an operation and one needs to find the optimal operation for each edge. Such a problem can be solved with discrete optimization algorithms. For example, by viewing the validation performance as the rewards, one can leveraged off-the-shelf reinforcement learning methods \cite{zophneural, bakerdesigning, zoph2018learning}. Moreover, evolutionary algorithms also achieve favorable performance for discrete NAS \cite{real2017large, xie2017genetic, real2019regularized}. 


\textbf{2) Efficient Searching Algorithms.}
The aforementioned NAS methods are able to find computationally more efficient network architectures than human design. However, their searching cost is a notable limitation, since their search procedure usually incorporates training many candidate networks from scratch to convergence to evaluate their validation accuracy. Motivated by this issue, a large number of works focus on developing low cost NAS algorithms. A basic idea in this direction is to reuse the previous candidates, \emph{e.g.}, adding/deleting layers \cite{cai2018efficient, elsken2018simple} and paths \cite{cai2018path} on top of currently found architectures or adopting existing architectures as network components \cite{liuhierarchical}.

Driven by these preliminary explorations, ENAS \cite{pham2018efficient} and DARTS \cite{liudarts} propose a parameter-sharing paradigm. They propose to construct a large computational graph that contains all possible connections and operations, such that each subgraph within it corresponds to a network architecture. The large graph is named as a \emph{super-net}, while all possible candidate networks share the same super-net parameters. Hence, one can train the super-net, and directly sample architectures from it without retraining any specific candidate network. The network selection process is usually formulated to be differentiable and accomplished efficiently via gradient-based optimization methods \cite{bender2018understanding, xiesnas, dong2019searching, zelaunderstanding, chen2022network}. Besides, some recent works focus on improving this procedure by introducing progressive searching mechanisms \cite{chen2019progressive, you2020greedynas}, introducing hyper-networks \cite{brocksmash, zhang2019graph} or training more proper super-nets for NAS \cite{guo2020single, yu2020bignas}.


\subsubsection{Latency-aware Neural Architecture Search}

From the lens of practical efficiency, an important challenge faced by NAS is the inference speed on real hardware (\emph{e.g.}, GPUs or CPUs). Since NAS usually leads to irregular network architectures, the obtained model with low theoretical computational cost may not be efficient in practice. To address this issue, recent NAS methods explicitly incorporate real latency into the optimization objective to achieve a good trade-off between real speed and accuracy \cite{caiproxylessnas, tan2019mnasnet, wu2019fbnet}. As representative examples, MobileNetV3 \cite{howard2019searching} leverages hardware-aware NAS to obtain the basic architecture, and modifies it manually. Once-for-all \cite{caionce} proposes to train a shared general super-nets, and perform NAS on top of it conditioned on the specific hardwares, yielding a state-of-the-art efficiency.

\subsection{Efficient Backbones for Video Understanding}
\label{sec:video_backbone}

In this subsection, we will focus on the efficient backbones for processing videos. Notably, videos consist of a series of frames, each of which is an image. In general, the aforementioned techniques for processing images are typically compatible with videos. Hence, here we mainly review the efficient modeling of the temporal relationships of video frames, including \emph{ConvNet-based} (Sec. 2.3.1) and \emph{Transformer-based} (Sec. 2.3.2) approaches.

\subsubsection{Efficient 3D ConvNets}

The most straightforward approach to modeling temporal relationships may be introducing 3D convolutional layers \cite{tran2015learning, carreira2017quo}, such that one can directly perform convolution in the space formed by frame height, width, and video duration. However, 3D convolution is computationally expensive, and many efficient backbones have been proposed to alleviate this problem.

\textbf{1) Marrying 2D and 3D Convolution.}
A basic idea is to avoid designing a pure 3D ConvNets, \emph{i.e.}, most of the feature extraction process may be accomplished by the efficient 2D convolution, while 3D convolution is only introduced at several particular positions. From the lens of macro-architecture, this goal can be attained by sequentially mixing 2D and 3D blocks, either first using 3D and later 2D or first 2D and later 3D \cite{zolfaghari2018eco, xie2018rethinking}. At the micro-architecture level, the group-wise or depth-width 3D convolution can be integrated in to the \emph{transform} module of 2D split-transform-merge architecture (Eq. (\ref{eq:split})) \cite{tran2019video, feichtenhofer2020x3d}.

\textbf{2) (2+1)D Networks.}
Another elegant idea is to decompose 3D convolution into two components: a 2D convolution that extract representation from video frames, and a temporal operation that only focuses on learning the temporal relationships. The former can directly adopt 2D neural operators, while the latter can be implemented using 1D temporal convolution \cite{tran2018closer, sun2015human, qiu2017learning}, adaptive 1D convolution \cite{liu2021tam}, and MLPs \cite{zhou2018temporal}.

\textbf{3) 2D Networks.}
In addition to the aforementioned approaches, the models with only 2D convolution may also be able to model temporal relationships. This is typically achieved by designing zero-parameter operations. For example, subtracting the features of adjacent frames to extract the motion information \cite{jiang2019stm, li2020tea}. The temporal-shift-based models \cite{lin2019tsm, zhang2021token, sudhakaran2020gate} propose to shift part of the channels of 2D features along the temporal dimension, performing information exchange among neighboring frames efficiently.

\textbf{4) Long/Short-term Separable Networks.}
Another important idea is modeling long/short-term temporal dynamics with separate network architectures. An representative work in this direction is SlowFast \cite{feichtenhofer2019slowfast}, which incorporate a lower temporal resolution slow pathway and a higher temporal resolution fast pathway. Many recent works \cite{li2020tea, wang2021tdn} further extend this idea.

\subsubsection{Transformer-based Video Backbones}

Driven by the success of ViTs \cite{dosovitskiy2021image}, a considerable number of recent works focus on facilitating efficient video understanding with self-attention-based models. In general, most of these works extend the aforementioned design ideas (including both image-based and video-based backbones) in the context of ViTs,  \emph{e.g.}, performing spatial-temporal local self-attention \cite{arnab2021vivit, bulat2021space, liu2022video}, combining self-attention and convolution \cite{liuniformer}, and performing 1D temporal attention in (2+1)D designs \cite{yan2022multiview, patrick2021keeping, bertasius2021space, neimark2021video}.



\subsection{Efficient Backbones for 3D Vision}
\label{sec:3d_backbone}

The perception and understanding of 3D scenes is not only a key ability of human intelligence, but also an important task for computer vision which are ubiquitous in real-world applications. In this subsection, we will review the backbones designed for processing 3D information efficiently. In general, the works in this direction can be categorized by the forms of model inputs, \emph{i.e.}, \emph{3D point clouds} (Sec. 2.4.1), \emph{3D voxels} (Sec. 2.4.2) and \emph{multi-view images} (Sec. 2.4.3).

\subsubsection{Point-based Models}

A fundamental type of 3D geometric data structure is the cloud of 3D points, where each point is represented by its three coordinates. PointNet \cite{qi2017pointnet} is the pioneering work that leveraging deep learning to process 3D point clouds. It adopts point-wise feature extraction with shared MLPs to maintain the permutation invariance. PointNet++ \cite{qi2017pointnet++} improves PointNet by facilitating capturing local geometric structures. On top of them, a number of works focus on how to aggregating local information effectively without increasing computational cost significantly. Representative approaches include introducing graph neural networks \cite{wang2019graph, wang2019dynamic}, projecting 3D points to regular grids to perform convolution \cite{xu2018spidercnn, tatarchenko2018tangent, thomas2019kpconv, li2018pointcnn}, aggregating the features of adjacent points using the weights determined by the local geometric structure \cite{liu2019relation, wu2019pointconv, liu2020closer}, and self-attention \cite{lai2022stratified, zhao2021point}. In particular, recent works have revealed that point-based models can achieve state-of-the-art computational efficiency with proper training and model scaling techniques \cite{qianpointnext}.

\subsubsection{Voxel-based Models}

The 3D point clouds can be further transformed to voxels, which are regular and can be directly processed with 3D convolution \cite{maturana2015voxnet}. Typically, the 3D space is divided into cubic voxel grids, while the features of the points in each grid will be averaged. The side length of the grid is named as the voxel resolution. An important technique for processing voxels efficiently is sparse convolution \cite{graham20183d, choy20194d, chen2022focal}, \emph{i.e.}, only performing convolution on the voxels with 3D points in them. Many works design backbone networks with this mechanism conditioned on the vision task of interest \cite{shi2020pv, shi2022pv, zhu2019class} for an optimal efficiency-accuracy trade-off. In addition, the point-based and voxel-based models can be combined to reduce the memory and computational cost \cite{liu2019point}. Some recent works have explored the automatic backbone design using NAS \cite{tang2020searching}

\subsubsection{Multi-view-based Models}

Multi-view projective analysis is another effective solution for understanding 3D shapes, where the 3D objects are projected into 2D images from varying visual angles and processed by 2D backbone networks \cite{su2015multi}. This idea can be implemented for recognition \cite{wei2020view, hamdi2021mvtn}, retrieval \cite{bai2016gift, jiang2019mlvcnn} and pose estimation \cite{kanezaki2018rotationnet}. An important challenge for these methods is how to fuse the multi-view features. Existing works have proposed to leverage LSTM \cite{jiang2019mlvcnn} or graph convolutional network \cite{wei2020view}.


\section{Dynamic Backbone Networks}
\label{sec:dy_backbone}

Although the advanced architectures introduced in Sec.~\ref{sec:backbone} have achieved significant progress in improving the inference efficiency of deep models, they generally have an intrinsic limitation: the computational graphs are kept the same during inference when processing different inputs with varying complexity. Such a \emph{static} inference paradigm inevitably brings redundant computation on some ``easy'' samples. To address this issue, dynamic neural networks \cite{han2021dynamic} have attracted great research interest in recent years due to their favorable efficiency, representation power, and adaptiveness \cite{han2021dynamic}.

Researchers have proposed various types of dynamic networks which can adapt their architectures/parameters to different inputs. Based on the granularity of adaptive inference, we categorize related works into \emph{sample-wise} (Sec.~\ref{sec:sample_wise}), \emph{spatial-wise} (Sec.~\ref{sec:spatial_wise}), and \emph{temporal-wise} (Sec.~\ref{sec:temporal_wise}) dynamic networks. Compared to the previous work \cite{han2021dynamic} which contains both vision and language models, we mainly focus on the computational efficient models for vision tasks in this survey. Moreover, more up-to-date works are included.

\subsection{Sample-wise Dynamic Networks}\label{sec:sample_wise}
The most common adaptive inference paradigm is processing each input sample (\emph{e.g.} an image) dynamically. There are mainly two lines of work in this direction: one aims at reducing the computation with decent network performance via dynamic \emph{architectures}, and the other adjusts network \emph{parameters} to boost the representation power with minor computational overhead. In this survey, we focus on the former line which typically reduces redundant computation for improving efficiency. Popular approaches include three types: 1) dynamic depth, 2) dynamic width, and 3) dynamic routing in a super network (SuperNet).

\begin{figure}[h]
    \begin{center}
        \includegraphics[width=0.6\linewidth]{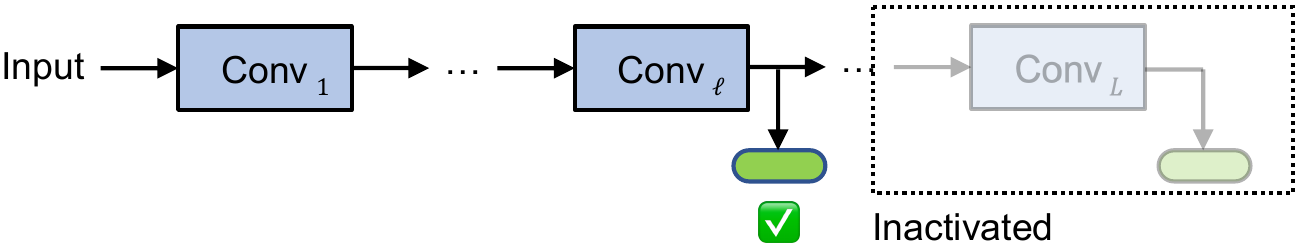}
    \end{center}
    \vskip -0.2in
    \captionsetup{font={small}}
    \caption{Dynamic early exiting. When the prediction at an early exit satisfies some criterion (the green tick), the inference procedure terminates, and the later computation will be skipped.}
    \label{fig:early_exit}
\end{figure}
\subsubsection{Dynamic Depth} \label{sec:dynamic_depth}
The inference procedure of a traditional (static) network can be written as
\begin{equation}
    \mathbf{y} = f(\mathbf{x}) = f^L\circ f^{L-1} \circ \dots \circ f^1(\mathbf{x}),
\end{equation}
where $f^{\ell},\ell=1,2,\dots,L$ is the $\ell$-th layer, and $L$ is the network depth. In contrast, networks with dynamic depth process each sample $\mathbf{x}_i$ with an adaptive number of layers:
\begin{equation}
    \mathbf{y}_i = f(\mathbf{x}) = f^{L_i}\circ f^{L_i-1} \circ \dots \circ f^1(\mathbf{x}), 
\end{equation}
where $1 \le L_i\le L$ is decided based on $\mathbf{x}_i$ itself.

There are mainly two common implementations to realize dynamic depth. The first is \emph{early exiting}, which means that the network predictions for some ``easy'' samples can be output at an intermediate layer without activating the deeper layers \cite{teerapittayanon2016branchynet,bolukbasi2017adaptive} (\figurename~\ref{fig:early_exit}). Researchers have found that multiple classifiers in a deep model may interfere with each other and degrade the performance by forcing early layers to capture semantic-level features \cite{huang2017multi}. To address this issue, multi-scale feature representation is adopted \cite{huang2017multi,yang2020resolution} to quickly produce coarse-scale features with rich semantic information. Instead of constructing intermediate exits in convolutional networks, the recent Dynamic Vision Transformer (DVT) \cite{wang2021not} realizes early exiting in cascaded vision Transformers which process images with different token numbers. Dynamic Perceiver \cite{han2023dynamic} proposes to integrate intermediate features and perform early-exit by introducing an addition attention-based path. Apart from architectural design, researchers have also proposed specialized techniques \cite{li2019improved,han2022learning} for training early-exiting models.

\begin{figure}[h]
    \begin{center}
        \includegraphics[width=0.45\linewidth]{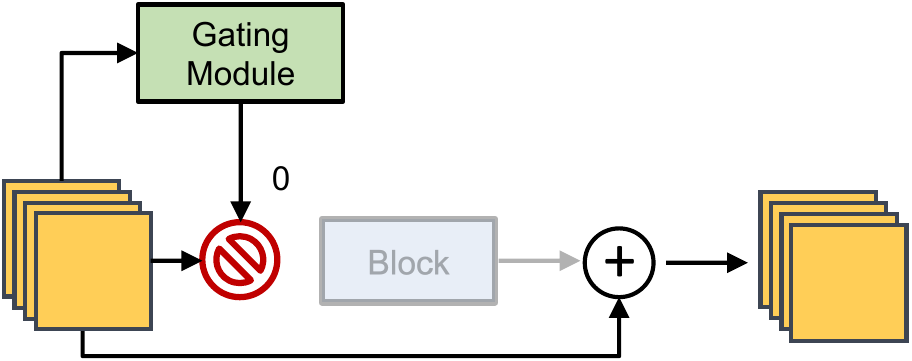}
    \end{center}
    \vskip -0.1in
    \captionsetup{font={small}}
    \caption{Dynamic layer skipping. A gating module is used to decide whether to execute the block.}
    \vskip 0.1in
    \label{fig:layer_skip}
\end{figure}

The aforementioned early-exiting methods dynamically terminate the forward propagation at a certain layer. An alternative approach to dynamic depth is \emph{skipping intermediate layers} in models with skip connection such as ResNets \cite{He_2016_CVPR} and vision Transformers \cite{dosovitskiy2021image} (\figurename~\ref{fig:layer_skip}). Let $\mathbf{x}^{\ell}$ and $f^{\ell}$ denote the feature and the computational unit at layer-$\ell$, a typical implementation of layer skipping is using a gating module $g^{\ell}(\cdot)$ to dynamically decide whether to execute $f^{\ell}$ \cite{wang2018skipnet,veit2018convolutional,meng2022adavit}: 
\begin{equation}
    \mathbf{x}^{\ell+1}=\mathbf{x}^{\ell} + g^{\ell}(\mathbf{x})\cdot f^{\ell}(\mathbf{x}), g^{\ell}(\mathbf{x})\in {0,1}.
\end{equation}

\begin{figure}[h]
    \begin{center}
        \includegraphics[width=0.45\linewidth]{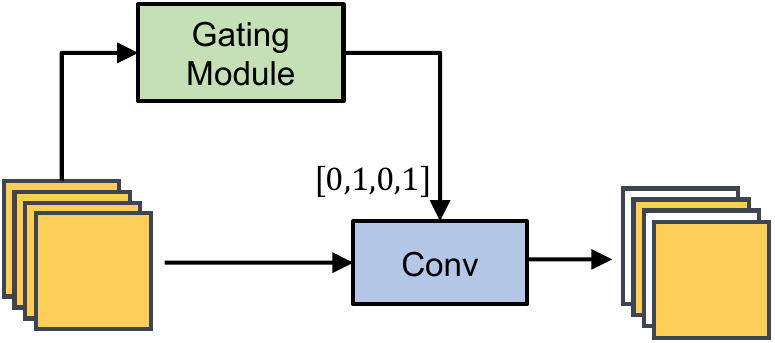}
    \end{center}
    \vskip -0.1in
    \captionsetup{font={small}}
    \caption{Dynamic channel skipping, which uses a gating module to decide the computation of convolution channels.}
    \label{fig:channel_skip}
\end{figure}

\subsubsection{Dynamic Width} 
Instead of skipping an entire layer, a less aggressive approach is adjusting the network \emph{width} to different inputs. In this direction, the most popular implementation is dynamically \emph{skipping the channels} in convolutional blocks via a gating module \cite{lin2017runtime,bejnordi2019batch,herrmann2020channel,li2021dynamic} (\figurename~\ref{fig:channel_skip}). Specifically, a gating module is first executed before conducting a convolution operation. The output of this gating module is a $C$-dimensional binary vector that decides whether to compute each channel, where $C$ is the output channel number. This implementation is similar to that in the aforementioned layer-skipping scheme. The most prominent difference is that the output of the gating module in layer skipping is a scalar, and the gating module in channel-skipping is required to output a vector controlling the computation of different channels. Apart from convolution layers, the same idea can also be applied in vision Transformers to dynamically skip channels in multi-layer perceptron (MLP) blocks \cite{meng2022adavit}.

\subsubsection{Dynamic Routing in SuperNets} 
Instead of skipping the computation of layers or channels in conventional network architectures, one can also realize data-dependent inference via \emph{dynamic routing} in super networks (SuperNets). A SuperNet usually contains various inference paths, and routing nodes are responsible for allocating each sample to the appropriate path. Let $\mathbf{x}_i^{\ell}$ denote the $i$-th node in layer-$\ell$, a general formulation of the computation for obtaining node-$j$ in the next layer can be written as
\begin{equation}
    \mathbf{x}_j^{\ell+1} = \sum_{i:\alpha_{i\rightarrow j}^{\ell} >0}  \alpha_{i\rightarrow j}^{\ell}f_{i\rightarrow j}^{\ell}(\mathbf{x}_i^{\ell}),
\end{equation}
where $f_{i\rightarrow j}^{\ell}$ is the transformation between node $i$ and $j$, and $\alpha_{i\rightarrow j}^{\ell}$ is the weight for this path which is calculated based on $\mathbf{x}_i^{\ell}$. If $\alpha_{i\rightarrow j}^{\ell}=0$, the transformation $f_{i\rightarrow j}^{\ell}$ can be skipped.

Extensive works have proposed different forms of SuperNets, such as tree structures \cite{yan2015hd,tanno2019adaptive,hazimeh2020tree}, dynamic mixture-of-experts \cite{yang2019condconv, pu2023adaptive}, and more general architectures \cite{cheng2020instanas,li2020learning}.

\subsection{Spatial-wise Dynamic Networks}\label{sec:spatial_wise}
It has been found that different spatial locations in an image contribute unequally to the performance of vision tasks \cite{zhou2016learning}. However, most existing deep models process different spatial locations with the same computation, leading to redundant computation on less important regions. To this end, spatial-wise dynamic networks are proposed to exploit the spatial redundancy in image data to achieve an improved efficiency. Based on the granularity of adaptive inference, we categorize relative works into pixel level, region level, and resolution level.

\subsubsection{Pixel-level Dynamic Networks}
A typical approach to spatial-wise adaptive inference is dynamically deciding whether to compute each pixel in a convolution block based on a binary mask \cite{dong2017more,verelst2020dynamic,xie2020spatially}. This form is similar to that in layer skipping and channel skipping (Sec.~\ref{sec:sample_wise}), except that the gating module is required to output a spatial mask. Each element of this spatial mask determines the computation of a feature pixel. In this way, the mask generators learn to locate the most discriminative regions in image features, and redundant computation on less informative pixels can be skipped.

The limitation of such pixel-level dynamic computation is that the acceleration is currently not supported by most deep learning libraries. The memory access cost can be heavier than static convolutions, and the computation parallelism is reduced due to sparse convolution. As a result, although the computation can be significantly reduced, the practical efficiency of these methods usually lags behind their theoretical efficiency. To this end, researchers have also proposed ``coarse-grained'' spatial-wise dynamic networks \cite{han2021spatially,hanlatency}, which means that an element of a spatial mask can decide a patch rather than a pixel. In this way, more contiguous memory access is realized for realistic speedup. Moreover, the scheduling strategies are also proven to have a considerable effect on the inference latency \cite{hanlatency}. It is also promising to co-design algorithm, scheduling, and hardware devices to better harvest the theoretical efficiency of spatial-wise dynamic networks.

Apart from skipping the computation of certain pixels, another line of work breaks the static reception field of traditional convolution and proposes deformable convolution \cite{dai2017deformable,zhu2019deformable,wang2022internimage}. Specifically, a lightweight module is used to learn the offsets for each feature pixel, and the convolution neighbors are sampled from arbitrary locations based on the predicted offsets. This idea has also been implemented in vision Transformers to enhance the performance of the local attention mechanism \cite{xia2022vision}.

\subsubsection{Region-level Dynamic Networks}
Instead of flexibly deciding which feature pixels to compute, another line of work aims at locating important regions (patches) in input images and cropping these patches for recognition tasks. For example, image recognition can be formulated as a sequential decision problem, in which an RNN is adopted to make predictions based on the cropped image patches \cite{mnih2014recurrent,li2017dynamic}. A multi-scale CNN with multiple sub-networks could also be used to perform the classification task based on cropped salient image patches \cite{fu2017look}. A lightweight module is placed between every two sub-networks to decide the coordinate and size of the salient patch.

Along this direction, the recent glance-and-focus network (GFNet) \cite{NeurIPS2020_7866, huang2022glance} proposes a general framework for region-level dynamic inference which is compatible with various visual backbones. It first ``glances'' a low-resolution input image, and then repeatedly ``focus'' on salient regions using reinforcement learning (RL) \cite{kaelbling1996reinforcement}. Moreover, early exiting (Sec.~\ref{sec:sample_wise}) is allowed, which means that the step number of ``focus'' can be dynamically adjusted for different input images.

\subsubsection{Resolution-level Dynamic Networks}
Most existing vision models process different images with the same resolution. However, the input complexity could vary, and not all images require a high-resolution representation. Ideally, low-resolution representations should be sufficient for those ``easy'' samples with large objects and canonical features. The early work \cite{hao2017scale} proposes to adaptively zoom input images in the face detection task. The recent resolution adaptive network (RANet) \cite{yang2020resolution} builds a multi-scale architecture, in which inputs are first processed with a low resolution and a small sub-network. Large sub-networks and high-resolution representations are conditionally activated based on early predictions. Instead of using a specialized structure, dynamic resolution network \cite{zhu2021dynamic} rescales each image with the resolution predicted by a small model and feeds the rescaled image to common CNNs.

Note that different spatial locations are still processed equally in the aforementioned methods. We categorize the relative works in this section since they mainly utilize the spatial redundancy of image inputs for efficient inference.

\subsection{Temporal-wise Dynamic Networks}\label{sec:temporal_wise}
As video data can be viewed as a sequence of image data, adaptive computation could also be performed along the temporal dimension due to the considerable redundancy in video  recognition tasks. Representative works can generally be divided into two lines: one processes video with recurrent models and dynamically save computation at certain time steps; the other aims at sampling key frames/clips and allocating the computation to these sampled frames.

\subsubsection{Dynamic Recurrent Models}
Different video frames are unequally informative. To this end, extensive studies propose to dynamically activate computation when updating the hidden state in recurrent models.  For example, LiteEval \cite{wu2019liteeval} establishes two different sized LSTM \cite{hochreiter1997long}. In each time step, a gating module is used to decide which LSTM should be executed for processing the current frame. AdaFuse \cite{meng2021adafuse} dynamically skips the computation of some convolution channels, and these channels are filled with the hidden state from the previous step. Moreover, the numerical precision \cite{sun2021dynamic} and image resolution \cite{meng2020ar} of different frames can also be dynamically decided.

The aforementioned works generally require a ConvNet for encoding each input frame before updating the hidden state. A more flexible solution is allowing the network to learn ``where to see''. In other words, networks can directly jump to an arbitrary temporal location in the video \cite{yeung2016end,alwassel2018action,wu2019adaframe} or perform early exiting \cite{fan2018watching,wu2020dynamic,ghodrati2021frameexit} instead of ``watch'' the entire video frame by frame.

\subsubsection{Dynamic Key Frame Sampling}
An alternative to skipping computation in recurrent networks is sampling key frames and then feeding the sampled frames rather than the whole video to a standard model. Reinforcement learning is a popular technique for training frame samplers \cite{tang2018deep,wu2019multi,zheng2020dynamic}.

A recent trend is simultaneously achieving dynamic inference from multiple perspectives. For example, AdaFocus and its variants \cite{wang2021adaptive,wang2022adafocus,wang2022adafocusv3,10155270} makes use of both spatial and temporal redundancy in video data. Dynamic architecture with 3D convolution \cite{li20212d} is also an interesting topic.

\section{Efficient Models for Downstream Computer Vision Tasks}
\label{sec:down_task}

In this section, we assume that a light-weighted backbone network has already been obtained, and discuss how to design task-specific heads or algorithms on top of them. The general aim is to facilitate accomplishing real-world computer vision tasks efficiently or even in real time. To this end, we will focus on three representative tasks, namely \emph{object detection} (Sec. \ref{sec:obj_det}), \emph{semantic segmentation} (Sec. \ref{sec:sem_seg}), and \emph{instance segmentation} (Sec. \ref{sec:ins_seg}), all of which have a strong need for accurate and real-time applications. Note that most of other more complex computer vision tasks (\emph{e.g.} visual object tracking) are mainly based on the three tasks we consider.

\subsection{Object Detection}
\label{sec:obj_det}

Object detection aims to answer two fundamental questions in computer vision: what visual objects are contained in the images, and where are them \cite{zou2023object}? The classification and localization results obtained by object detection usually serve as the basis of other vision tasks, \emph{e.g.}, instance segmentation, image captioning, and object tracking. The algorithms for object detection can be roughly categorized into \emph{two-stage} (Sec. 4.1.1) and \emph{one-stage} (Sec. 4.1.2). In the following, we will discuss them respectively from the lens of computational efficiency.

\subsubsection{Two-stage Detectors}

Object detection with deep learning starts from the two stage paradigm. The pioneer work, RCNN \cite{girshick2014rich, girshick2015region}, proposes to first crop a set of object proposals from the images, and classify them with deep networks. On top of it, SPPNet \cite{he2015spatial} avoids repeatedly inferring the backbones by adaptively pooling the features of the regions of interest. Fast RCNN \cite{girshick2015fast} simultaneously train a detector and a bounding box regressor in the same network, leading to more than 200 times of speedup than RCNN. Faster R-CNN \cite{ren2015faster, ren2017faster} and its improvements \cite{dai2016r, li2017light} introduce a region proposal network that cheaply generates object proposals from the features, yielding the first nearly real-time deep learning detector. The feature pyramid networks further propose to leverage the feature maps at varying scales to detect the object with different sizes respectively, which improves the detection accuracy significantly without sacrificing the efficiency \cite{lin2017feature}.

\subsubsection{One-stage Detectors}

The major motivation behind the two-stage detects is the ``coarse-to-fine'' refining, \emph{i.e.}, first obtaining the coarse proposals, and then refining the localization and discrimination results on top of these proposals, such that an excellent detection performance can be achieved. Despite the aforementioned techniques proposed to improve the efficiency of this procedure, the speed and the complexity of two-stage detectors are usually not applicable to real-time applications. In contrast, the one-stage detectors directly output the detection results in a single step, yielding much faster inference speed with a decent accuracy.

\textbf{1) Bounding-box-based Methods.}
The first deep-learning-based one-stage detector is YOLO \cite{redmon2016you}. YOLO divides the image into grid regions and simultaneously predicts the bounding boxes and the classification results conditioned on each region. The subsequent works of YOLO \cite{redmon2018yolov3, bochkovskiy2020yolov4, redmon2017yolo9000, wang2021you, wang2022yolov7} focus on further improving the localization performance or classification accuracy without affecting the practical speed. The latest version, YOLOv7 \cite{wang2022yolov7}, achieves a state-of-the-art effectiveness-efficiency trade-off.

In addition to YOLO, SSD \cite{liu2016ssd} improves the accuracy of one-stage detectors by detecting the objects at different scales on different layers of the network. RetinaNet \cite{lin2017focal} proposes a focal loss to encourage the model to focus more on the difficult, misclassified examples, which boosting the accuracy of one-stage detectors effectively.

\textbf{2) Point-based Methods.}
The aforementioned detection methods mostly learn to produce the ground-truth bounding boxes on top of pre-defined anchor boxes. Despite the effectiveness, this paradigm suffers from a lot of design hyper-parameters and an imbalance between positive/negative boxes during training. To address this issue, CornerNet \cite{law2018cornernet} proposes to directly predict the top-left corner and bottom-right corner of candidate boxes. Many subsequent works extend this point-based setting. For example, FCOS \cite{tian2019fcos} predicts the distances from each location in feature maps to the four sides of the bounding box. ExtremeNet \cite{zhou2019bottom} learns to detect the extreme points the center of bounding boxes. CenterNet \cite{zhou2019objects} further considers each object to be a single center point and regresses all the attributes (2D/3D size, orientation, depth, locations, etc.) based on this point.

\textbf{3) Transformer-based Methods.}
In recent years, N. Carion \emph{et al.} propose an end-to-end Transformer-based detection network, DETR \cite{carion2020end}. DETR views detection as a set prediction problem, where the results are obtained based on several object queries. Deformable DETR \cite{zhu2021deformable} addresses the long convergence issue of DETR by introducing a deformable mechanism to self-attention.

\subsection{Semantic Segmentation}
\label{sec:sem_seg}

The aim of semantic segmentation is to predict the semantic label of each pixels \cite{minaee2021image, holder2022efficient}, \emph{e.g.}, if a pixel belongs to a car, a bike, etc. Here we summarize existing efficient semantic segmentation methodologies based on their paradigms \emph{i.e.}, \emph{encoder-decoder} (Sec. 4.2.1), \emph{multi-branch} (Sec. 4.2.2) and others (Sec. 4.2.3).

\subsubsection{Encoder-decoder}
A popular approach is to first extract the low-resolution discriminative representations with a multi-stage backbone network, up-sample the deep features to the input resolution with a decoder, and then produce the pixel-wise predictions. This procedure is named as ``encoder-decoder'' \cite{zhao2017pyramid, badrinarayanan2017segnet}. To improve the efficiency of this paradigm, many works propose to  design light-weighted decoders. Representative methods include introducing  split-transform-merge architectures \cite{paszke2016enet, chaurasia2017linknet, mehta2018espnet, mehta2019espnetv2, guo2022segnext} (Eq. (\ref{eq:split})), developing efficient approximations of the computationally intensive dilated convolution \cite{romera2017erfnet, li2019dabnet, lo2019efficient}, and introducing dense connections \cite{lo2019efficient, krevso2020efficient}. In addition, it is efficient to simultaneously feed the low-level and high-level features into the decoder, \emph{i.e.}, comprehensively leveraging both of them improves the accuracy without introducing notable computational overhead \cite{chen2018encoder, poudel2019fast, zhuang2019shelfnet, orsic2019defense, krevso2020efficient}.

\subsubsection{Multi-branch Models}
Another popular efficient paradigm is designing multi-branch architectures. Typically, the model consist of two types of paths: 1) context paths with low-resolution feature maps and large receptive fields, aiming to extract discriminative information; and 2) spatial paths that preserve the low-level spatial information. These paths are fused in a parallel \cite{yu2018bisenet, chen2018encoder, poudel2018contextnet, yu2021bisenet, fan2021rethinking} or cascade \cite{zhao2018icnet} fashion, yielding high-resolution but semantically rich deep representations for segmentation.

\subsubsection{Others}
In recent years, some new ideas have been proposed to facilitate efficient semantic segmentation. For example, processing deep features with self-attention layers \cite{wang2019lednet, li2019dfanet, hu2020real}, designing segmentation models with NAS \cite{shaw2019squeezenas, liu2019auto, chenfasterseg}, adjusting the architecture of the decoder conditioned on the inputs \cite{li2020learning}. More recently, a considerable number of papers seek to design efficient semantic segmentation models on top of ViTs \cite{cheng2021per, zheng2021rethinking, xie2021segformer, zhang2022semantic, cheng2022masked}. These works mainly focus on achieving a state-of-the-art performance with as less computational cost as possible.

\subsection{Instance Segmentation}
\label{sec:ins_seg}

Instance segmentation can be seen as a combination of object detection and semantic segmentation, where the model needs to detect the instances of objects, demarcate their boundaries and recognize their categories \cite{minaee2021image, zhang2021survey}. Existing works in this direction can be categorized into \emph{two-stage} (Sec. 4.3.1) and \emph{End-to-end} (Sec. 4.3.2).

\subsubsection{Two-stage Approaches}
From the lens of efficiency, a notable milestone of deep-learning-based instance segmentation is the proposing of Mask R-CNN \cite{he2017mask}. Mask R-CNN is developed by introducing mask segmentation branches on the basis of Faster R-CNN \cite{ren2015faster}. It enjoys high computational efficiency by directly obtaining the regions of interest from the feature maps. In contrast, MaskLab \cite{chen2018masklab} improved Faster R-CNN by adding the semantic segmentation and direction prediction paths. To improve the accuracy of Mask R-CNN, MS R-CNN \cite{huang2019mask} predicts the quality of the predicted instance masks and prioritizes more accurate mask predictions during validation. PANet \cite{liu2018path} introduces a path augmentation mechanism to facilitate the bottom-up information interaction of feature maps. HTC \cite{chen2019hybrid} proposes a hybrid task cascade framework to learn more discriminative features progressively while integrating complementary features in the meantime.

\subsubsection{End-to-end Approaches}
Another liner of works focus on realizing efficient end-to-end instance segmentation. SOLO \cite{wang2020solo, wang2020solov2} achieves this by introducing the ``instance categories'', which assigns categories to each pixel within an instance according to the instance's location and size, thus converting instance segmentation into a pure dense classification problem. YOLACT \cite{bolya2019yolact} and BlendMask \cite{chen2020blendmask} propose to first generate a set of prototype masks, and then combines them with per-instance mask coefficients or attention scores. Inspired by SSD \cite{liu2016ssd} and RetinaNet \cite{lin2017focal}, TensorMask \cite{chen2019tensormask} build an efficient sliding-window-based instance segmentation framework.

\section{Model Compression Techniques}
\label{sec:compress}


Deep networks necessitate substantial resources, including energy, processing capacity, and storage. These resource requirements diminish the suitability of deep networks for resource-constrained devices \cite{akmandor2018smart}. Furthermore, the extensive resource requirements of deep networks become a bottleneck for real-time inference and executing deep networks on browser-based applications. To address these drawbacks of deep networks, various model compression techniques have been proposed in existing literature. Several comprehensive reviews on model compression techniques exist \cite{mishra2020survey, cheng2017survey}. These reviews categorize model compression techniques, discuss challenges, provide overviews, solutions, and future directions of model compression techniques. We adopt their classification structure but place a greater emphasis on vision-related works. Specifically, we categorize existing research into network pruning \cite{xu2020convolutional,liang2021pruning}, network quantization \cite{gholamisurvey,liang2021pruning}, low-rank decomposition \cite{denton2014exploiting,jaderberg2014speeding}, knowledge distillation \cite{hinton2015distilling,romero2014fitnets,peng2019correlation}, and other techniques \cite{larsen2020weight,kim2016pvanet}. For readers interested in a particular category, we recommend consulting these more targeted reviews \cite{xu2020convolutional,liang2021pruning,gholamisurvey,gou2021knowledge}.

\begin{figure}[h]
    \begin{center}
        \includegraphics[width=0.55\linewidth]{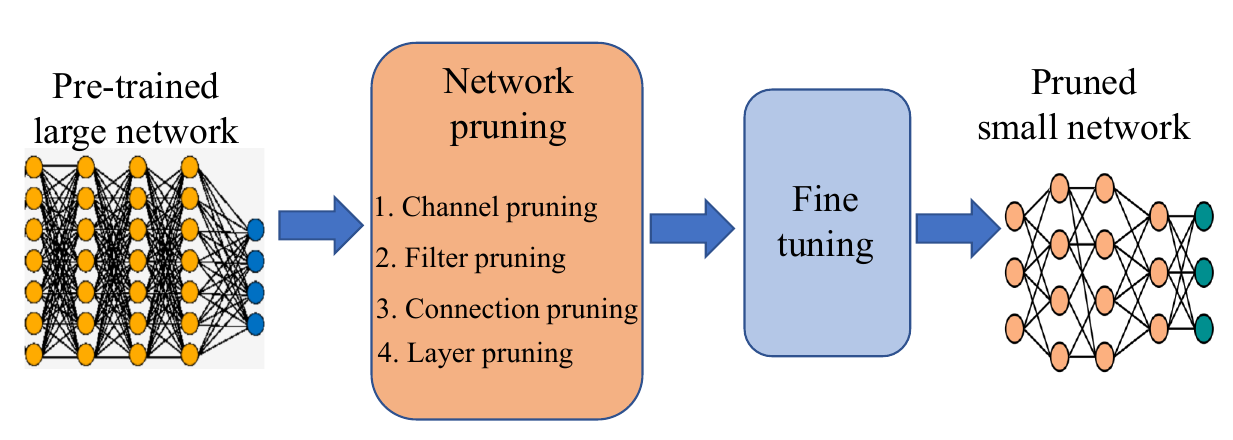}
    \end{center}
    \vskip -0.1in
    \captionsetup{font={small}}
    \caption{The steps of network pruning.}
    \label{fig:prune}
\end{figure}

\subsection{Network Pruning}\label{sec:network_pruning}

Network pruning is one of the most prevalent techniques for reducing the size of a deep learning model by eliminating inadequate components, such as channels, filters, neurons, or layers, resulting in a light-weighted model. Network pruning techniques can be categorized into four types: channel pruning, filter pruning, connection pruning, and layer pruning. These techniques help decrease the storage and computation requirements of deep networks. A typical pruning algorithm consists of two stages: evaluating and pruning unimportant parameters, followed by fine-tuning the pruned model to restore accuracy. The steps and categories are illustrated in Figure \ref{fig:prune}.

In deep networks, the inputs provided to each layer are channeled. Channel pruning involves removing unimportant channels to reduce computation and storage requirements. Various channel pruning schemes have been proposed \cite{he2017channel,liu2017learning, zhao2019variational}. Convolutional operations in ConvNets incorporate a large number of filters to enhance performance. Increases in filter quantities result in a significant growth in the number of floating-point operations. Filter pruning eliminates unimportant filters, thus reducing computation \cite{luo2017thinet,huynh2017deepmon,denton2014exploiting}. The number of input and output connections to a layer in deep networks determines the number of parameters. These parameters can be used to estimate the storage and computation requirements of deep networks. Connection pruning is a direct approach to reduce parameters by removing unimportant connections \cite{han2016eie,louizos2017bayesian,parashar2017scnn}. Layer pruning involves selecting and deleting certain unimportant layers from the network, leading to ultra-high compression of the deep network. This is particularly useful for deploying deep networks on resource-constrained computing devices, where ultra-high compression is necessary. Some layer pruning approaches have been proposed to substantially reduce both storage and computation requirements \cite{tan2019mnasnet,he2018multi}. However, layer pruning may result in a higher accuracy compromise due to the structural deterioration of deep networks.

\subsection{Network Quantization}\label{sec:network_quantization}
Network quantization aims to compress the original network by reducing the storage requirements of weights. It can be categorized into linear quantization and nonlinear quantization. Linear quantization focuses on minimizing the number of bits needed to represent each weight, while nonlinear quantization involves dividing weights into several groups, with each group sharing a single weight.


\subsubsection{Linear Quantization} Utilizing 32-bit floating-point numbers to represent weights consumes a substantial amount of resources. Consequently, linear quantization employs low-bit number representation to approximate each weight. Suyog \emph{et al.} contend that the weights of deep networks can be represented by 16-bit fixed-point numbers without significantly reducing classification accuracy \cite{gupta2015deep}. Some studies further compress ConvNets to 8-bit \cite{gysel2016hardware,mathew2017sparse}. In the extreme case of a 1-bit representation for each weight, binary weight neural networks emerge. The primary concept is to directly learn binary weights or activation during model training. Several works directly train ConvNets with binary weights, including BinaryConnect \cite{courbariaux2015binaryconnect}, BinaryNet\cite{courbariaux2016binarized}, and XNOR \cite{rastegari2016xnor}.


\subsubsection{Nonlinear Quantization} Nonlinear quantization entails dividing weights into several groups, with each group sharing a single weight. Gong \emph{et al.} initially employ the k-means algorithm to cluster weight parameters and replace the parameter values with the clustering center values, substantially reducing the network's storage space \cite{gong2014compressing}. Wu \emph{et al.} further quantize convolution filters, fully connected layers, and other parameters \cite{wu2016quantized}. Chen \emph{et al.} randomly assign weights to hash buckets, with each hash bucket sharing a single weight \cite{chen2015compressing}. Han \emph{et al.} combine network pruning, parameter quantization, and Huffman coding to achieve significant reductions in storage and memory \cite{han2015deep}.

\subsection{Knowledge Distillation}\label{sec:knowledge_distillation}
\begin{figure}[h]
    \begin{center}
        \includegraphics[width=0.55\linewidth]{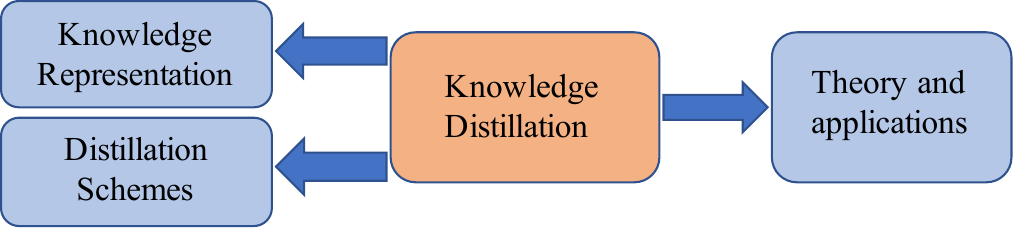}
    \end{center}
    \vskip -0.1in
    \captionsetup{font={small}}
    \caption{Knowledge distillation. The section mainly contains knowledge representation, distillation schemes, theory and applications.}
    \label{fig:kd}
\end{figure}

Knowledge distillation (KD) \cite{hinton2015distilling} is a widely adopted technique for transferring ``dark knowledge'' from a high-capacity model (teacher) to a more compact model (student) in order to achieve various types of efficiency. The two primary aspects of KD are knowledge representation and distillation schemes. In this section, we concentrate on existing research in these two technical areas and further summarize the theoretical exploration and application progress of KD in computer vision, as illustrated in Figure \ref{fig:kd}.

\subsubsection{Knowledge Representation} \label{sec:knowledge_representation}

Drawing on \cite{gou2021knowledge}, we examine different forms of knowledge in the following categories: response-based knowledge, feature-based knowledge, and relation-based knowledge. Response-based knowledge typically refers to the neural response of the teacher model's final output layer, with the main idea being to directly emulate the teacher model's final prediction. The most prevalent response-based knowledge for image classification is soft targets \cite{hinton2015distilling}. In object detection tasks, the response may include logits along with the bounding box offset \cite{chen2017learning}. For semantic landmark localization tasks, such as human pose estimation, the teacher model's response may consist of a heatmap for each landmark \cite{zhang2019fast}.

Feature-based knowledge pertains to the feature representation derived from intermediate layers. Fitnets \cite{romero2014fitnets} are the first to introduce intermediate representations, which subsequently inspire the development of various methods \cite{zagoruyko2016paying,mun2018learning,kim2018paraphrasing,jin2019knowledge,xu2020feature,wang2020exclusivity,wang2021towards,wang2022learn}. Relation-based knowledge further investigates the relationships between different feature layers \cite{yim2017gift,lee2018self,passalis2020heterogeneous} or data samples \cite{liu2019knowledge,tung2019similarity,park2019relational,peng2019correlation}. For instance, Yim \emph{et al.} \cite{yim2017gift} propose calculating the relations between pairs of feature maps using the Gram matrix, while Liu \emph{et al.} \cite{liu2019knowledge} suggest transferring the instance relationship graph, which defines instance features and relationships as vertices and edges, respectively.

\subsubsection{Distillation Schemes} \label{sec:distillation_schemes}


The learning schemes of knowledge distillation can be classified into three main categories based on the synchronization of the teacher model's update with the student model: offline distillation, online distillation, and self-distillation.


In offline distillation, the teacher model is usually assumed to be pre-trained. The primary focus of offline methods is to enhance various aspects of knowledge transfer, including knowledge representation and the design of loss functions. Vanilla knowledge distillation \cite{hinton2015distilling} serves as a classic example of offline distillation methods. Most prior knowledge distillation methods operate in an offline manner.


In cases where a high-capacity, high-performance teacher model is unavailable, online distillation provides an alternative. In this approach, both the teacher model and the student model are updated simultaneously, allowing for an end-to-end trainable knowledge distillation framework. Deep mutual learning \cite{zhang2018deep} introduced a method for training multiple neural networks collaboratively, where any given network can serve as the student model while the others act as teachers. Numerous online knowledge distillation methods have been proposed \cite{zhu2018knowledge,guo2020online,walawalkar2020online}, with multi-branch architecture \cite{zhu2018knowledge} and ensemble techniques \cite{guo2020online,chen2020online} being widely adopted.



Self-distillation refers to a learning process in which the student model acquires knowledge independently, without the presence of teacher models, whether pre-trained or virtual. Several studies have explored this idea in various contexts. For instance, Zhang \emph{et al.} \cite{zhang2019your} propose a method for distilling knowledge from deeper layers to shallower ones for image classification tasks. Similarly, Hou \emph{et al.} \cite{hou2019learning} employ attention maps from deeper layers as distillation targets for lower layers in object detection tasks. In contrast, Yang \emph{et al.} \cite{yang2019snapshot} introduce snapshot distillation, where checkpoints from earlier epochs are considered as teachers to distill knowledge for the models in later epochs. Additionally, Wang \emph{et al.} \cite{wang2021towards} suggest constraining the outputs of the backbone network using target class activation maps.

\subsubsection{Theory and Applications} \label{sec:theory_application}


 A wide range of knowledge distillation methods has been extensively employed in vision applications. Initially, most knowledge distillation methods were developed for image classification \cite{hinton2015distilling,yim2017gift,peng2019few,wangefficient,wang2022tc3kd} and later extended to other vision tasks, including face recognition \cite{feng2020triplet,wang2020exclusivity}, action recognition \cite{garcia2018modality,stroud2020d3d}, object detection \cite{chen2017learning,deng2019relation,dai2021general}, semantic segmentation \cite{liu2019structured,jiao2019geometry,wang2020intra,hou2020inter}, depth estimation \cite{guo2018learning,xu2018pad,tosi2019learning,pilzer2019refine,pirvu2021depth,wang2021knowledge}, image retrieval \cite{chen2021feature,jang2022deep}, video captioning \cite{zhang2020object,pan2020spatio,dong2021revisiting}, and video classification \cite{zhang2018better,bhardwaj2019efficient}, among others.

Despite the significant practical success, relatively few works have focused on the theoretical or empirical understanding of knowledge distillation \cite{yuan2019revisit,tang2020understanding,cheng2020explaining,wangefficient}. Hinton \emph{et al.} \cite{hinton2015distilling} suggest that the success of KD could be attributed to learning similarities between categories. Yuan \emph{et al.} \cite{yuan2019revisit} posited that dark knowledge not only encompasses category similarities but also imposes regularization on student training. They indicate that KD is a learned label smoothing regularization (LSR). Tang \emph{et al.} \cite{tang2020understanding} propose approach where, in addition to regularization and class relationships, another type of knowledge, instance-specific knowledge, is also used by the teacher to rescale the student model's per-instance gradients. Chen \emph{et al.} \cite{cheng2020explaining} quantify the extraction of visual concepts from the intermediate layers of a deep learning model to explain knowledge distillation. Wang \emph{et al.} \cite{wangefficient} connect KD with the information bottleneck and empirically validate that preserving more mutual information between feature representation and input is more important than improving the teacher model's accuracy. Overall, theoretical research remains limited compared to the diverse and numerous applications.

\subsection{Low-rank Factorization}
\label{sec:lowrank_factorization}


 Convolution kernels can be viewed as 3D tensors. Ideas based on tensor decomposition are derived from the intuition that there is structural sparsity in the 3D tensor. In the case of fully connected layers, they can be viewed as 2D matrices (or 3D tensors), and low-rankness can also be helpful. The key idea of low-rank factorization is to find an approximate low-rank tensor that is close to the real tensor and easy to decompose. Low-rank factorization is beneficial for both tensors and matrices.


There are several typical low-rank methods for compressing 3D convolutional layers. Lebedev \emph{et al.} \cite{lebedev2015speeding} propose Canonical Polyadic (CP) decomposition for kernel tensors. They use nonlinear least squares to compute the CP decomposition for a better low-rank approximation. Since low-rank tensor decomposition is a non-convex problem and generally difficult to compute, Jaderberg \emph{et al.} use iterative schemes to obtain an approximate local solution \cite{jaderberg2014speeding}. Then, Tai \emph{et al.} find that the particular form of low-rank decomposition in \cite{jaderberg2014speeding} has an exact closed-form solution, which is the global optimum, and present a method for training low-rank constrained ConvNets from scratch \cite{tai2016convolutional}.


Many classical works have exploited low-rankness in fully connected layers. Denil \emph{et al.} reduce the number of dynamic parameters in deep models using the low-rank method \cite{denil2013predicting}. Zhang \emph{et al.} introduce a Tucker decomposition model to compress weight tensors in fully connected layers \cite{zhang2017tucker}. Lu \emph{et al.} adopt truncated singular value decomposition to decompose the fully connected layer for designing compact multi-task deep learning architectures \cite{lu2017fully}. Sainath \emph{et al.} explore a low-rank matrix factorization of the final weight layer in deep networks for acoustic modeling \cite{sainath2013low}.

\subsection{Hybrid Techniques}\label{sec:hybrid_techniques}

Apart from the four categories of mainstream techniques mentioned above, there are other techniques for network compression. Some studies have attempted to integrate orthogonal techniques to achieve more significant performance \cite{han2015deep,tung2018clip,polinomodel}. Some works have designed compact networks \cite{howard2017mobilenets,ma2018shufflenet,han2020ghostnet} or efficient convolutions \cite{huang2018condensenet,chollet2017xception}, which have been discussed in Sec. \ref{sec:backbone}.

\section{Efficient Deployment on Hardware}
\label{sec:hardware}

The aforementioned works mostly design network architectures based on their theoretical computation (\emph{e.g.} floating operations, FLOPs). However, there is often a gap between theoretical computation and practical latency on hardware devices \cite{sandler2018mobilenetv2,ma2018shufflenet}. Realistic efficiency can be influenced by other factors such as hardware properties and scheduling strategies. Along this direction, we review relative works from the following perspectives: 1) hardware-aware neural architecture search (Sec.~\ref{sec:hardware_friendly}); 2) acceleration software libraries and hardware design (Sec.~\ref{sec:acceleration_tools}); and 3) algorithm-software codesign techniques.

\subsection{Hardware-aware Model Design}\label{sec:hardware_friendly}
As the practical latency of models can be influenced by many factors other than theoretical computation, the commonly used FLOPs is an inaccurate proxy for network efficiency. Ideally, one should develop efficient models based on specific hardware properties. However, hand-designing networks for different hardware devices can be laborious. Therefore, automatically \emph{searching} for efficient architectures is emerging as a promising direction. Compared to the traditional NAS methods \cite{zophneural,liu2018darts}, this line of works can generate appropriate models which satisfy different hardware constraints and gain realistic efficiency in practice. For example, ProxylessNAS \cite{caiproxylessnas} establishes a latency prediction function based on realistic tests on targeted hardware, and the predicted latency is then directly used as a regularization item in the NAS objective. A similar idea is also implemented by MnasNet \cite{tan2019mnasnet} to search for efficient models on mobile devices. The following works FBNet \cite{wu2019fbnet}, FBNet-v2 \cite{wan2020fbnetv2} and OFA \cite{Cai2020Once-for-All} have improved NAS techniques. 

Apart from the traditional static models, the hardware-aware design paradigm has also been applied to develop \emph{spatial-wise dynamic networks} (Sec.~\ref{sec:spatial_wise}) \cite{hanlatency}.

Note that we mainly give a brief introduction of basic ideas in this work due to the page limit. For more detailed techniques we refer the readers to the survey \cite{chitty2022neural} which specifically focuses on this topic.

\subsection{Acceleration Tools}\label{sec:acceleration_tools}
In addition to architectural design, the efficient deployment of algorithms on hardwares also requires acceleration software libraries or specific hardware accelerators.

\textbf{1) Software Libraries.} 
Extensive efforts have been made to accelerate model inference on different hardware platforms. For example, NVIDIA TensorRT \cite{vanholder2016efficient} is widely used to deploy models for optimized inference on GPUs. NNPACK ({https://github.com/Maratyszcza/NNPACK}.), CoreML \cite{thakkar2019introduction} and TinyEngine \cite{lin2020mcunet} are representative tools on multi-core CPUs, Apple silicons, and microcontrollers (MCUs), respectively. Cross-platform tools such as Tencent TNN ({https://github.com/Tencent/TNN}). and Apache TVM \cite{chen2018tvm} have also emerged as popular development tools.

\textbf{2) Hardware Accelerators.}
Apart from adapting neural architectures to given hardware devices, another line of works studies accelerators from the hardware perspective to enable fast inference of deep models. For example, DianNao \cite{chen2014diannao} focuses on memory behavior and proposes an accelerator that simultaneously improves the inference speed and energy consumption of deep models. An FPGA-based accelerator is proposed quantitatively analyze the throughput of CNNs with the help of the classical roofline model \cite{williams2009roofline}. In addition to the regular deep networks, researchers have also proposed accelerators to improve the inference efficiency of spatially sparse convolution \cite{albericio2016cnvlutin,8735526}.

\subsection{Algorithm-Hardware Co-design}
The aforementioned methods typically improve the inference efficiency from the perspective of either algorithm or hardware. Ideally, one should expect algorithms and hardware can ``cooperate'' with each other to further push forward the Pareto frontier between accuracy and efficiency trade-off. Along this direction, extensive efforts have been made based on the highly flexible and versatile Field Programmable Gate Arrays (FPGA) platform, and NAS techniques (Sec.~\ref{sec:hardware_friendly}) are widely used to search for hardware-friendly network structures \cite{zhang2019neural,hao2019fpga,jiang2020standing,jiang2020hardware,abdelfattah2020best}. The recent MCUNet series \cite{lin2020mcunet,lin2021mcunetv2,lin2022ondevice} has enabled both inference and training on MCUs based on algorithm-hardware co-design with the help of their proposed tiny-Engine tool (Sec.~\ref{sec:acceleration_tools}).

The co-designing method has also been applied to the field of dynamic neural networks, especially for efficient spatially adaptive convolution \cite{hua2019boosting,song2020drq,colleman2021processor} and attention \cite{zhou2022energon} operations.

\section{Challenges and Future Directions}
\label{sec:challenge}


Despite the significant advances in the field of computationally efficient deep learning in recent years, numerous open challenges warrant further research. In this section, we summarize these challenges and discuss potential future directions.

\subsection{Designing General-purpose Backbones}

The efficient extraction of discriminative representations from raw inputs has been established as a critical cornerstone for practical deep learning applications, as demonstrated in the existing literature. Light-weighted backbone networks are commonly employed to achieve this goal. As a result, a significant challenge lies in the design of efficient, general-purpose backbones. Potential avenues of investigation in this area encompass enhancing current convolution and self-attention operators via manual design \cite{howard2017mobilenets, liu2021swin}, employing automated architecture search methodologies \cite{caionce}, and amalgamating these approaches to create comprehensive efficient modules \cite{tangghostnetv2}. Specifically, the exploration of innovative information aggregation methods beyond convolution and self-attention appears promising, for instance, clustering algorithms \cite{ma2023image}, LSTM \cite{tatsunami2022sequencer}, and graph convolution \cite{han2022vision}. Moreover, an emerging area of interest involves enabling backbone networks to accommodate multi-modal inputs (\emph{e.g.}, text, images, and videos) and execute multiple visual tasks (\emph{e.g.}, retrieval, classification, and visual question answering) \cite{wang2022image, yuan2021florence}. Consequently, the development of mobile-level multi-modal and multi-task visual foundation models could present an intriguing direction for future research.

\subsection{Developing Task-specialized Models}

In addition to the architectural advancements in backbone models, tailoring deep learning methodologies to specific computer vision tasks of interest has been demonstrated as crucial. Two research challenges of particular significance in this domain can be identified. Firstly, the exploitation of representations extracted by backbones to efficiently obtain task-specific features is essential, for example, multi-scale features for object detection and multi-path fused features for semantic segmentation. A potential solution to this challenge could involve designing specialized, efficient decoders (\emph{e.g.}, utilizing NAS \cite{ghiasi2019fpn, chenfasterseg}). Secondly, it is important to streamline the multi-stage design of visual tasks (\emph{e.g.}, two-stage object detection \cite{ren2017faster} and instance segmentation \cite{he2017mask} algorithms) to achieve end-to-end paradigms with minimal performance compromises. Additionally, the removal of time-consuming components, such as non-maximum suppression (NMS) \cite{zou2023object}, is crucial. A promising area for future research may involve the development of an efficient, unified, and end-to-end learnable interface for a majority of prevalent computer vision tasks \cite{chen2022a}.

\subsection{Deep Networks for Edge Computing}


In practical applications, extant research predominantly focuses on conventional hardware, such as GPUs and CPUs. However, within the realm of edge computing, there is an increasing demand for the deployment of deep learning models on Internet of Things (IoT) devices and microcontrollers. These diminutive devices are characterized by their minimal size, low power consumption, affordability, and ubiquity \cite{lin2020mcunet}. The development of deep learning algorithms specifically adapted for such devices represents an exigent research direction. MCUNets \cite{lin2020mcunet, lin2021mcunetv2, lin2022ondevice} have provided an initial exploration by optimizing the design, inference, and training of ConvNets for these devices. Another prospective concept involves the creation of spiking neural networks \cite{tavanaei2019deep}, which, when co-designed with hardware, can yield energy-efficient solutions.


\subsection{Leveraging Large-scale Training Data}

Contemporary large visual backbone models have exhibited remarkable scalability in response to the increasing volumes of training data \cite{dosovitskiy2021image}, that is, the model's performance consistently enhances as more training data becomes accessible. However, it is generally arduous for computationally efficient models with a reduced number of parameters to capitalize on this high-data regime to the same extent as their larger counterparts. For example, the improvements attained by pre-training light-weighted models on expansive ImageNet-22K/JFT datasets are typically inferior to those observed in larger models \cite{dosovitskiy2021image, liu2021swin, liu2022convnet}. This challenge is similarly experienced by self-supervised learning algorithms, where the methods effective for larger models frequently produce limited gains for smaller models \cite{he2020momentum, he2022masked}. As a result, a propitious avenue of research involves the exploration of effective scalable supervised and unsupervised learning algorithms for light-weighted models, allowing them to reap the benefits of an unlimited amount of data without incurring the expense of acquiring annotations. Some recent works on novel training algorithms have started to preliminarily explore this direction \cite{wang2021revisiting, tan2021efficientnetv2, li2022autoprog, Ni2022Incub, wang2023efficienttrain}.

\subsection{Practical Efficiency}

While numerous extant studies have attained low theoretical computational costs, they may be hindered by the restricted practical efficiency. For example, certain irregular network architectures discovered through NAS may display considerable latency on GPUs/CPUs, and the models employing group or depth-wise convolution may exhibit reduced gains in actual speedup relative to their theoretical computational efficiency. To tackle this challenge, researchers might consider integrating the speed on practical hardwares as an objective in architecture design \cite{tan2019mnasnet, caionce} or utilizing efficient implementation software \cite{vanholder2016efficient, chen2018tvm}. From a hardware design standpoint, one potential direction involves the creation of model-specialized hardware platforms \cite{zhang2019neural, hao2019fpga, jiang2020standing, jiang2020hardware}.

\subsection{Model Compression Approaches}

Network compression algorithms, encompassing network pruning, quantization, and knowledge distillation, have exhibited a robust capacity to diminish the inference costs associated with deep networks. However, several avenues of investigation remain unexplored. For instance, while the overarching concept of model compression is not confined to a particular vision task, a majority of algorithms predominantly concentrate on image classification, rendering their extension to other tasks non-trivial. A significant research direction entails the development of general-purpose, task-agnostic model compression techniques. Furthermore, strategies such as network pruning may yield irregular architectural topologies, potentially impairing the practical efficiency of deep learning models. Consequently, the examination of practically efficient compression methodologies constitutes a propitious area for future research.



















\section*{Acknowledgments}
This work is supported in part by the National Key R\&D Program of China (2021ZD0140407), the National Natural Science Foundation of China (62022048, 62276150), Guoqiang Institute of Tsinghua University and Beijing Academy of Artificial Intelligence.

\bibliographystyle{unsrt}
\bibliography{reference.bib}

\end{document}